\pdfoutput=1
\documentclass[letterpaper]{article}
\usepackage{aaai2027}
\usepackage[hyphens]{url}
\usepackage{graphicx}
\usepackage{booktabs}
\usepackage{multirow}
\usepackage{amsmath,amssymb}
\usepackage{algorithm}
\usepackage[noend]{algpseudocode}
\usepackage{microtype}
\usepackage{enumitem}
\usepackage{tikz}
\usetikzlibrary{positioning,arrows.meta,fit,shapes.geometric}
\usepackage{natbib}
\newcommand{\resultval}[1]{#1}

\newcommand{\BroadInventory}{\resultval{96.2}}
\newcommand{\BroadMasked}{\resultval{88.7}}
\newcommand{\BroadFOne}{\resultval{79.4}}
\newcommand{\BroadSemEq}{\resultval{77.8}}
\newcommand{\BroadPages}{\resultval{91.2}}
\newcommand{\TypedPrec}{\resultval{86.0}}
\newcommand{\TypedRec}{\resultval{73.7}}
\newcommand{\DenseFOne}{\resultval{88.2}}
\newcommand{\DenseSemEq}{\resultval{89.1}}
\newcommand{\ClosureFireRate}{\resultval{93}}

\newcommand{\NRequests}{\resultval{1{,}000}}
\newcommand{\PlannerFeas}{\resultval{100.0}}
\newcommand{\PlannerUtility}{\resultval{0.868}}
\newcommand{\WeightedTimeouts}{\resultval{40}}
\newcommand{\EETCoverage}{\resultval{99.5}}
\newcommand{\EETFeas}{\resultval{99.5}}
\newcommand{\EETUtility}{\resultval{0.853}}

\newcommand{\PlannerUtilityGainBest}{\resultval{+0.066}}  
\newcommand{\InvGainBest}{\resultval{+3.5}}               
\newcommand{\MaskedGainBest}{\resultval{+8.9}}            
\newcommand{\FOneGainBest}{\resultval{+4.3}}              
\newcommand{\SemEqGainBest}{\resultval{+5.1}}             
\newcommand{\PageSavingExh}{\resultval{47}}               
\newcommand{\DenseFOneDelta}{\resultval{8.8}}             
\newcommand{\DenseSemEqDelta}{\resultval{11.3}}           
\newcommand{\GapShrink}{\resultval{68}}                   

\newcommand{\CIExtract}{\resultval{[95.3, 96.9]}}
\newcommand{\CIPlanner}{\resultval{[0.841, 0.889]}}
\newcommand{\CIEET}{\resultval{[0.822, 0.879]}}
\newcommand{\SignFlipExtract}{\resultval{$p<0.001$}}
\newcommand{\SignFlipPlanner}{\resultval{$p<0.001$}}
\newcommand{\ExtractWins}{\resultval{81}}                 

\newcommand{\sig}{$^{\dagger}$}

\newcommand{\sigSchemaScore}{\sig}
\newcommand{\sigRefClosure}{\sig}
\newcommand{\sigFixedPoint}{\sig}
\newcommand{\sigAdapters}{\sig}
\newcommand{\sigFallback}{\sig}

\newcommand{\sigDiffusion}{\sig}
\newcommand{\sigUnlock}{\sig}
\newcommand{\sigClarify}{\sig}
\newcommand{\sigWeighted}{\sig}
\newcommand{\sigGEApplic}{\sig}

\title{KnowPlan: From Heterogeneous University Catalogs to Personalized Degree Plans}

\author{
Shuheng Cao$^{1,*}$,
Weijia Zhang$^{2,*}$,
Jiaqi Wu$^{1,*}$,
Xiyun Hu$^{3,\dagger}$,
Yat Yang$^{1,\dagger}$,
Juqy Chen$^{1,\dagger}$,
Zhaoxiang Feng$^{1,\dagger}$
}

\affiliations{
\begin{tabular}{@{}c@{}}
$^{1}$University of California, San Diego
\quad
$^{2}$Yale University
\\[0.2em]
$^{3}$University of North Carolina, Chapel Hill
\\[0.45em]
{\small
$^{*}$Co-first authors
\qquad
$^{\dagger}$Equal contribution
}
\end{tabular}
}

\begin{document}
\maketitle
\begin{abstract}Planning degree pathways from public sources remains difficult because institutional curricula are distributed across heterogeneous webpages, JSON endpoints, and PDF formats, while valid student-specific plans must satisfy prerequisite logic, overlapping degree requirements, course availability, workload limits, and individual preferences. We present \textsc{KnowPlan}, an extraction-first framework that separates curriculum reconstruction from personalized degree planning and explicitly evaluates the interface between the two stages. To prevent student-specific planning objectives from biasing source discovery, we introduce \emph{CatalogBrowse}, a web  exploration agent that selects legal browsing actions using lower-confidence estimates of expected marginal coverage over a finite set of atomic catalog obligations per unit of source access. CatalogBrowse combines deterministic platform adapters with a span-constrained clause-to-AST model fallback and terminates using a closure certificate over index, schema, provenance, and reference completeness rather than a learned reward threshold. It exposes the reconstructed curriculum through an exact contract of three provenance-linked JSON documents. We further introduce \emph{DegreeMap}, which consumes only these documents, compiles them into a typed requirement hypergraph, and performs lexicographic CP-SAT optimization over hard feasibility, completion horizon, workload and scheduling risk, personalized utility, and future option value. Each objective is optimized within the proven optimum of the preceding stage, preserving certifiability under a fixed solver budget. Across a 100-university broad evaluation and a six-university dense evaluation, CatalogBrowse achieves \BroadInventory \% inventory recall and \BroadMasked \% masked-source recovery while using \PageSavingExh \% fewer source accesses than exhaustive crawling. DegreeMap maintains \PlannerFeas\% hard feasibility and improves personalized utility by \PlannerUtilityGainBest{} over the strongest baseline. The complete pipeline certifies \EETCoverage\% of planning requests and attains a utility gap of only 0.015 relative to planning over a privileged gold curriculum graph.
\end{abstract}
\begin{figure*}[t]
\centering
\includegraphics[width=0.9\textwidth]{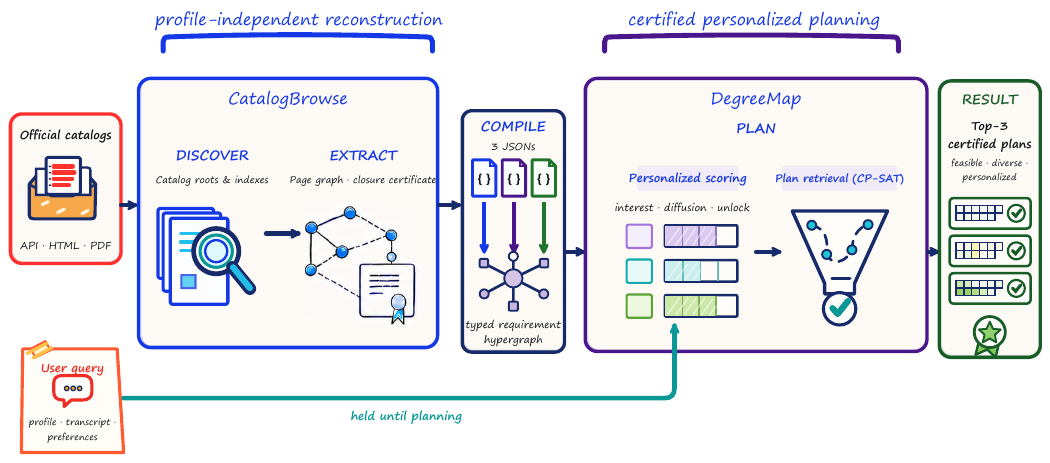}
\caption{KnowPlan is extraction-first. CatalogBrowse compiles an institution into three provenance-linked JSON documents before DegreeMap receives a user profile.}
\label{fig:system}
\end{figure*}

\section{Introduction}
Academic degree pathway planning is a constraint-driven decision problem.
Students must sequence courses to satisfy prerequisite logic, degree and
general-education requirements, unit limits, and term-level offerings while
balancing workload and individual goals. Prior work has studied course
recommendation, personalized sequencing, and long-horizon curriculum
optimization
\citep{parameswaran2011recommendation,xu2016personalized,mohamed2015decision}.
Recent LLM systems and educational knowledge graphs further support
natural-language advising and curriculum retrieval
\citep{vandeventer2024interests,spahicbogdanovic2025imodulebuddy,
hogan2021knowledge,abusaliha2024systematic}.
These approaches, however, largely assume that a sufficiently complete
curriculum representation is already available.

Constructing that representation from public university sources is itself
difficult. Graduation rules are distributed across catalogs, departmental
pages, program handbooks, college websites, JSON endpoints, and PDFs with no
shared schema. They contain nested \texttt{all-of}, \texttt{any-of}, and
\texttt{choose-$k$} clauses, requirement overlaps, applicability conditions,
and policy exceptions. More importantly, allowing a personalized planner to
control source acquisition creates \emph{acquisition--planning circularity}.
The current student objective determines which pages are opened, the resulting
partial graph determines which plans appear feasible, and the same graph is
then used to validate the plan. A missing rule remains invisible because it was
never acquired.

This gap is not addressed by either neighboring literature. Web-agent
benchmarks reward completing a specified task or locating a sufficient answer
\citep{deng2023mind2web,zhou2024webarena,he2024webvoyager}, while LLM-based
planners generally begin from a supplied domain representation
\citep{liu2023llmp,yuan2025llmap}. Reliable degree planning instead requires
profile-independent curriculum acquisition before student-specific
optimization.

We present \textsc{KnowPlan}, an extraction-first framework that separates
these stages and measures their interface. \textsc{CatalogBrowse} receives a
target institution but no user profile. It explores a frozen official-source
snapshot by estimating marginal coverage over a finite set of atomic catalog
obligations per unit of source access. It combines deterministic platform
adapters with a span-constrained clause-to-AST fallback and stops only when a
closure certificate holds over index, schema, provenance, and reference
completeness. It emits exactly three provenance-linked JSON documents for
courses and typed prerequisites, program requirements, and general-education
frameworks. \textsc{DegreeMap} consumes only these documents, compiles them
into a typed requirement hypergraph, and performs lexicographic CP-SAT
optimization over hard feasibility, completion horizon, workload and
scheduling risk, personalized utility, and future option value. Each objective
is optimized within the preceding objective's proven optimum, preserving
certifiability within the solver budget.

We evaluate \textsc{KnowPlan} on a 100-university broad track and a six-school
dense track. A masked-source test removes a stratified $10\%$ of index pages and
all direct links to them, exposing entities that task-directed crawlers would
otherwise miss. A paired planning track runs identical profiles on privileged
gold graphs and extracted graphs to isolate acquisition loss.
\textsc{CatalogBrowse} reaches \BroadInventory\% inventory recall and
\BroadMasked\% masked-source recovery while using \PageSavingExh\% fewer source
accesses than an exhaustive crawler. \textsc{DegreeMap} maintains
\PlannerFeas\% exact feasibility, improves personalized utility by
\PlannerUtilityGainBest{} over the strongest baseline, and reduces the
completion horizon from $6.7$ to $6.3$ terms. End to end, \textsc{KnowPlan}
certifies \EETCoverage\% of requests with a utility gap of $0.015$ relative to
the privileged gold graph.

Our contributions are:
\begin{enumerate}
    \item We formulate degree planning from public sources as a two-stage
    acquisition-and-optimization problem and make
    acquisition--planning circularity directly measurable.

    \item We introduce \textsc{CatalogBrowse}, a profile-independent agent
    with obligation-driven exploration, provenance-linked extraction, and
    certificate-based stopping.

    \item We introduce \textsc{DegreeMap}, which combines a typed requirement
    hypergraph with exact lexicographic optimization, and evaluate the full
    pipeline through masked-source and paired gold-versus-extracted tracks.
\end{enumerate}

\section{Related Work}
\paragraph{Web agents that stop when the task is answered.}
Benchmarks for autonomous web agents score whether an agent completes a specified task. Mind2Web \citep{deng2023mind2web}, WebArena \citep{zhou2024webarena}, WebVoyager \citep{he2024webvoyager}, BrowserGym \citep{chezelles2024browsergym}, and Mind2Web~2 \citep{gou2025mindtwo} all score task completion in a rendered or sandboxed environment. Related work extends agent evaluation and reasoning to data questions, capability-conditioned navigation, and multimodal affordances, while agent-first data systems emphasize infrastructure designed around such agents \citep{ma2026can,su2026capnav,wang2026affordancer1,liu2025supporting}. In the web benchmarks above, reward is defined by a task instance, so an agent holding a sufficient answer may stop, and nothing charges it for what it never looked at. Our masked-source track charges exactly that. Go-Browse \citep{gandhi2025gobrowse} is closest in treating a site as a graph to explore structurally rather than traverse once, and our ablations isolate what a schema ledger, a reference-closure objective, and a fixed-point certificate add on top of that view. ReAct \citep{yao2023react} supplies the loop our ReAct-Qwen baseline follows, and API access beats rendering where endpoints exist \citep{song2024beyondbrowsing}, which motivates our per-platform adapters.

\paragraph{Planners that are handed the domain.}
LLM+P \citep{liu2023llmp} formalizes a natural-language problem and delegates optimization to an exact planner, LLM-Modulo \citep{gundawar2024llmmodulo} keeps the model in a generate-and-critique loop around external verifiers, and LLMAP \citep{yuan2025llmap} performs multi-objective route planning under preferences. TravelPlanner \citep{xie2024travelplanner} shows language agents remain weak at hard constraints even when the domain is handed to them. We inherit the separation of formalization from optimization, but the domain is not given. It is the output of the extraction stage, which is why end-to-end feasibility is verified against a gold graph rather than the extracted graph the plan was built from. Verifying against the extracted graph would reward an extractor for dropping the constraint it failed to parse, and that inversion is the circularity this paper is about.

\paragraph{Prerequisite extraction, at a different unit.}
Recovering prerequisite structure from educational text is established. Liang et al.\ \citep{liang2017recovering} recover concept prerequisites from course dependencies, and LectureBank \citep{li2019lecturebank} and TutorialBank \citep{fabbri2018tutorialbank} provide labeled corpora. These target \emph{concept}-level relations over curated text. Our unit is an institution's official course-level rule, the text must first be located across a heterogeneous public ecosystem rather than supplied, and the output must be an executable Boolean rule carrying provenance rather than a binary label, because it is consumed by an exact solver that will certify a plan against whatever it is handed.

\paragraph{Degree planning, and the assumption we remove.}
Course recommendation subject to graduation requirements has been posed as recommendation with complex constraints \citep{parameswaran2011course}, a line of work that assumes the requirement structure is already encoded. Removing that assumption is the contribution. Acquiring the structure under a completion certificate and then planning over it exactly converts a modeling assumption into a measured quantity, and Table~\ref{tab:e2e} reports what it was worth: 0.015 of utility under our best configuration and up to 0.117 under a weaker extractor.

\section{Method}
\subsection{Problem Formulation}
For institution $s$ and catalog year $y$, let $\mathcal{W}_{s,y}$ be a frozen official-source environment. A source action reveals pages, links, sections, or structured endpoints. The extractor must produce
$J_s=(J_s^{\mathrm{course}},J_s^{\mathrm{program}},J_s^{\mathrm{GE}})$,
where each fact is linked to exact provenance. A deterministic compiler maps $J_s$ to a typed requirement hypergraph $G_s$.

A user request $q$ specifies a target program, completed courses, preferences, and possibly unknown preference dimensions. The planner outputs a ranked set of term-indexed plans $\Pi(q,G_s)$. A certified plan satisfies every supported hard constraint, and among certified plans the planner optimizes graduation time, load and risk, personalized utility, and option value lexicographically.

\subsection{CatalogBrowse: School-Level Catalog Exploration}
\begin{figure*}[t]
\centering
\includegraphics[width=0.68\textwidth]{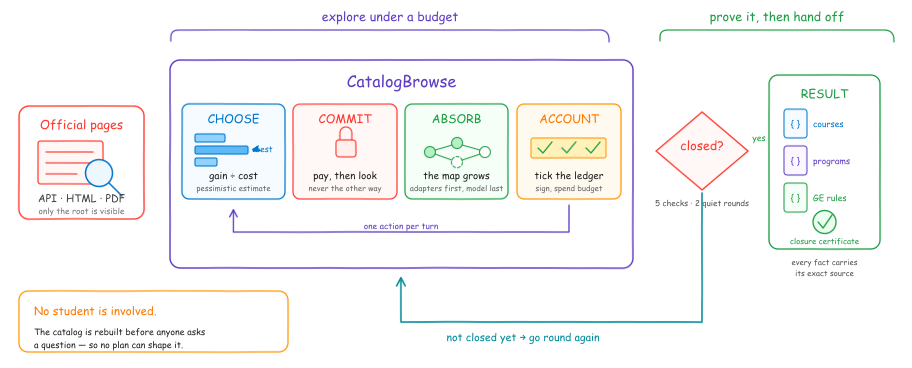}
\caption{CatalogBrowse. The agent starts able to see only a root URL. Each turn it scores every legal action by lower-confidence expected obligation gain per unit of source access, seals and pays for one action \emph{before} the page is revealed, folds the result into its page graph, and ticks a five-kind obligation ledger. It leaves the loop only when the closure certificate holds, and emits exactly three provenance-linked JSON documents. No student profile is visible at any point in this stage.}
\label{fig:catalogbrowse}
\end{figure*}

\begin{figure*}[t]
\centering
\includegraphics[width=0.68\textwidth]{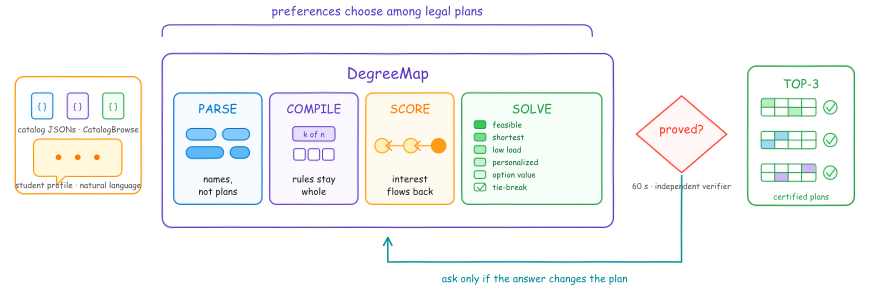}
\caption{DegreeMap. It reads only the three documents CatalogBrowse produced plus the student's own request, keeps compound requirements such as choose-two-of-three as single objects rather than expanding them into pairwise clauses, and propagates interest backward along prerequisite paths so a gateway course inherits the value of what it unlocks. Objectives are then solved as a lexicographic ladder in which legality is settled first and only what survives each rung competes on the next, so a preference can select among certified plans but never in place of them.}
\label{fig:degreemap}
\end{figure*}

\begin{table*}[t]
\centering
\small
\caption{Broad-100 extraction profile.}
\label{tab:extract}
\begin{tabular}{lrrrrr}
\toprule
Method & Inventory Recall $\uparrow$ & Masked Recovery $\uparrow$ & Typed F1 $\uparrow$ & AST SemEq $\uparrow$ & Pages/School $\downarrow$\\
\midrule
Static BFS & \resultval{82.4} & \resultval{64.8} & \resultval{69.8} & \resultval{65.1} & \resultval{118.0}\\
Platform-Exhaustive & \resultval{92.7} & \resultval{72.6} & \resultval{73.6} & \resultval{70.5} & \resultval{171.4}\\
Go-Browse-style & \resultval{90.8} & \resultval{79.8} & \resultval{75.1} & \resultval{72.7} & \resultval{132.6}\\
ReAct-Qwen & \resultval{87.9} & \resultval{76.2} & \resultval{74.0} & \resultval{70.8} & \resultval{109.8}\\
\textbf{CatalogBrowse} & \textbf{\BroadInventory} & \textbf{\resultval{88.7}} & \textbf{\BroadFOne} & \textbf{\BroadSemEq} & \textbf{\resultval{91.2}}\\
\bottomrule
\end{tabular}
\end{table*}

\begin{table}[t]
\centering
\small
\caption{Broad and dense tracks on the two columns they share. Dense-6 is fully annotated over five source ecosystems. The gap is the cost of the long tail of platform conventions that no adapter anticipates, and it is the reason we report the broad track as the operating point.}
\label{tab:dense}
\begin{tabular}{lrrr}
\toprule
Track & Schools & Typed F1 $\uparrow$ & AST SemEq $\uparrow$\\
\midrule
Broad-100 & 100 & \BroadFOne & \BroadSemEq\\
Dense-6 & 6 & \DenseFOne & \DenseSemEq\\
\midrule
Difference & & \DenseFOneDelta & \DenseSemEqDelta\\
\bottomrule
\end{tabular}
\end{table}

\begin{table*}[t]
\centering
\small
\caption{Gold-graph personalized planning profile. Feasibility is evaluated independently from personalization utility.}
\label{tab:planner}
\begin{tabular}{lrrrrrr}
\toprule
Method & Hard Feas. $\uparrow$ & Major $\uparrow$ & GE $\uparrow$ & Prereq. $\uparrow$ & Utility $\uparrow$ & Terms $\downarrow$\\
\midrule
Direct Qwen & \resultval{73.5} & \resultval{82.1} & \resultval{76.8} & \resultval{88.6} & \resultval{0.736} & \resultval{7.1}\\
Requirement-only CP-SAT & \resultval{100.0} & \resultval{100.0} & \resultval{100.0} & \resultval{100.0} & \resultval{0.746} & \resultval{6.3}\\
LLM+P & \resultval{98.5} & \resultval{99.0} & \resultval{98.5} & \resultval{99.0} & \resultval{0.784} & \resultval{6.5}\\
LLMAP-style MSGS & \resultval{96.5} & \resultval{97.5} & \resultval{96.0} & \resultval{98.0} & \resultval{0.802} & \resultval{6.7}\\
\textbf{DegreeMap} & \textbf{\PlannerFeas} & \textbf{\resultval{100.0}} & \textbf{\resultval{100.0}} & \textbf{\resultval{100.0}} & \textbf{\PlannerUtility} & \textbf{\resultval{6.3}}\\
\bottomrule
\end{tabular}
\end{table*}

\begin{table}[t]
\centering
\scriptsize
\setlength{\tabcolsep}{3.5pt}
\caption{Source-to-plan result profile. The gold graph is a privileged upper reference. Changing the extractor primarily shifts coverage and feasibility, while changing the planner produces the larger utility shift, both swaps also affect the other columns, and neither component alone recovers the full system. Both of the first two columns are over all \NRequests{} pairs, and feasibility is checked against the gold hypergraph rather than the extracted one.}
\label{tab:e2e}
\begin{tabular}{lrrrr}
\toprule
System & Cover.\ $\uparrow$ & Feas.\ $\uparrow$ & Util.\ $\uparrow$ & Gap $\downarrow$\\
\midrule
Gold + DegreeMap & \resultval{100.0} & \resultval{100.0} & \resultval{0.868} & \resultval{0.000}\\
Static BFS + DegreeMap & \resultval{89.5} & \resultval{88.1} & \resultval{0.751} & \resultval{0.117}\\
Go-Browse + DegreeMap & \resultval{95.0} & \resultval{94.7} & \resultval{0.821} & \resultval{0.047}\\
CatalogBrowse + CP-SAT & \resultval{99.0} & \resultval{98.5} & \resultval{0.761} & \resultval{0.107}\\
\textbf{CatalogBrowse + DegreeMap} & \textbf{\EETCoverage} & \textbf{\EETFeas} & \textbf{\EETUtility} & \textbf{\resultval{0.015}}\\
\bottomrule
\end{tabular}
\end{table}

\begin{table*}[t]
\centering
\scriptsize
\caption{Two ablation studies. A dagger marks a variant whose paired clustered contrast against the full system excludes zero at 95\%. The two rows carrying the argument are $-$~fixed point, which lowers cost and loses 4.4 points of inventory recall, and Weighted sum, which raises utility above the full system while losing 4.0 points of certified feasibility to solver-budget exhaustion rather than to a relaxed constraint.}
\label{tab:ablations}
\begin{minipage}[t]{0.49\textwidth}
\centering
\textbf{CatalogBrowse ablation}\\[2pt]
\begin{tabular}{lrrrr}
\toprule
Variant & Inv. & F1 & SemEq & Pages\\
\midrule
Full & \resultval{96.2} & \resultval{79.4} & \resultval{77.8} & \resultval{91.2}\\
$-$ schema score\sigSchemaScore & \resultval{92.1} & \resultval{75.8} & \resultval{73.9} & \resultval{102.7}\\
$-$ reference closure\sigRefClosure & \resultval{94.3} & \resultval{76.1} & \resultval{74.0} & \resultval{87.6}\\
$-$ fixed point\sigFixedPoint & \resultval{91.8} & \resultval{75.4} & \resultval{73.5} & \resultval{79.3}\\
$-$ adapters\sigAdapters & \resultval{84.6} & \resultval{70.2} & \resultval{67.4} & \resultval{110.8}\\
$-$ Qwen fallback\sigFallback & \resultval{96.0} & \resultval{72.8} & \resultval{67.1} & \resultval{88.4}\\
\bottomrule
\end{tabular}
\end{minipage}\hfill
\begin{minipage}[t]{0.49\textwidth}
\centering
\textbf{DegreeMap ablation}\\[2pt]
\begin{tabular}{lrrr}
\toprule
Variant & Feas. & Utility & Rel. resp.\\
\midrule
Full & \resultval{100.0} & \resultval{0.868} & \resultval{100.0}\\
$-$ diffusion\sigDiffusion & \resultval{100.0} & \resultval{0.832} & \resultval{91.5}\\
$-$ unlock\sigUnlock & \resultval{100.0} & \resultval{0.839} & \resultval{93.0}\\
$-$ clarification\sigClarify & \resultval{98.5} & \resultval{0.850} & \resultval{67.5}\\
Weighted sum\sigWeighted & \resultval{96.0} & \resultval{0.873} & \resultval{100.0}\\
$-$ GE applicability\sigGEApplic & \resultval{92.5} & \resultval{0.860} & \resultval{100.0}\\
\bottomrule
\end{tabular}
\end{minipage}
\end{table*}

\paragraph{State and observability.}
CatalogBrowse maintains a dynamic page graph, a discovered-but-unopened frontier, a platform adapter, index ledgers, parsed entities, unresolved references, missing fields, budgets, and a hash-chained trace (Figure~\ref{fig:catalogbrowse}). At initialization the agent sees only root metadata. Page content is revealed only after an action has been precommitted, and newly revealed links expand the legal action frontier. Precommitment is enforced by the environment rather than by convention, so an agent cannot inspect a page in order to decide whether to pay for it. Algorithm~\ref{alg:catalogbrowse} gives the loop.

\paragraph{Atomic obligations.}
Loosely defined completion signals are why browsing agents stop early: any weighted sum of them is maximized by opening whichever pages are easiest to score on. We instead fix a finite set of atomic obligations $\mathcal{O}$. An obligation is discovering an official index entry, instantiating an entity, filling a required field, resolving a typed reference, or attaching provenance. For partial state $\psi$ the weighted completion potential is
\begin{equation}
F(\psi)=\sum_{o\in\mathcal{O}}v_o\mathbf{1}[o\text{ is completed in }\psi].
\end{equation}
The policy selects the legal action with maximal lower-confidence expected marginal obligation gain per realized cost:
\begin{equation}
a_t=\arg\max_{a\in\mathcal{A}(\psi_t)}
\frac{\operatorname{LCB}\!\left(\mathbb{E}[F(\psi_{t+1})-F(\psi_t)\mid a,\psi_t]\right)}{c(a)}.
\end{equation}
The lower bound is a conservatism device, not an exploration bonus: an action kind with few observed outcomes gets a wide interval and a discounted score, so the policy prefers evidence it has already validated. What pays for an unfamiliar action is the obligation itself, which keeps generating positive expected gain until it is bound or explicitly failed.

\paragraph{What is and is not guaranteed.}
The policy carries no optimality bound. The greedy guarantee of \citet{golovin2011adaptive} needs a ground set fixed in advance, actions whose coverage depends only on their own target, and scoring by true conditional expectation, and we have none of the three: $\mathcal{O}$ grows as parsing reveals structure, multi-page evidence makes some actions complementary, and equation~(2) scores by a lower bound. What is guaranteed is a property of the output.

\textit{Proposition 1.} If the closure certificate holds for a run over snapshot $\mathcal{W}_{s,y}$, then every fact in the three emitted documents carries an in-bounds span of a page the agent actually opened, every discovered index entry is either bound to an entity or recorded as explicitly failed, every schema-required field is bound or explicitly unresolved, and each of these is recomputable from the hash-chained trace by a third party who never runs the agent. The certificate locates evidence, it does not adjudicate it.

\paragraph{Deterministic-first parsing.}
API, DOM, PDF, and structured-record adapters extract indexes, identifiers, titles, units, pagination, and simple relations. The model is invoked only against a visible exact span and a finite clause-to-AST schema. It cannot emit URLs, course identifiers, relation types, or actions, because the identifier enum offered to it is built from tokens literally present in the span, and returned evidence is re-located in the page text before it is accepted. An output that fails validation is retried once with validator feedback, and a second failure stays unresolved.

\paragraph{Stopping rule.}
Closure is a hard terminal certificate rather than a reward term. CatalogBrowse stops only after every discovered index entry and cursor has been processed or explicitly failed, two consecutive closure passes add no page, entity, or reference, every required field is present or explicitly unresolved, every provenance offset lies in bounds on a page the agent opened, and the three documents validate. The certificate is snapshot-relative and makes no claim about the live Web, which is what makes it checkable. A third party holding the same frozen snapshot can recompute every condition from the hash-chained trace. Reruns use deterministic policy settings and a fixed model-serving configuration, while the receipt chain exposes any trace divergence rather than assuming byte-identical model output.

\begin{algorithm}[t]
\caption{CatalogBrowse}
\label{alg:catalogbrowse}
\small
\begin{algorithmic}[1]
\Require Frozen source environment $\mathcal{E}$, roots $\mathcal{R}_0$, obligations $\mathcal{O}$, budget $B$
\Ensure Three JSON documents and closure certificate
\State Initialize page graph $G$, frontier $Q\gets\mathcal{R}_0$, entity store $V$, completed obligations $Y$, unresolved set $U$, trace $T$
\While{$B>0$}
  \If{\Call{ClosureSatisfied}{$G,Q,V,U,Y$}} \State \textbf{break} \EndIf
  \State $A\gets\Call{LegalActions}{G,Q,V,U,B}$
  \ForAll{$a\in A$}
    \State $s(a)\gets\Call{ScoreAction}{a,G,V,U,Y,\mathcal{O}}$ \Comment{$\operatorname{LCB}\big(\sum_{o\notin Y}v_o\widehat p(o\mid a,G,V)\big)/c(a)$}
  \EndFor
  \State $a^*\gets\arg\max_{a\in A}s(a)$ using deterministic tie-breaking
  \State $r\gets\Call{Precommit}{a^*,G,Q,T}$
  \State $z\gets\mathcal{E}.\Call{Execute}{a^*,r}$
  \State $(G,Q)\gets\Call{UpdatePageGraph}{G,Q,z}$
  \State $\widehat V\gets\Call{ParseObservation}{z}$
  \State $V\gets\Call{NormalizeAndMerge}{V,\widehat V}$
  \State $(V,U)\gets\Call{ResolveReferences}{V}$
  \State $(Y,\mathcal{O})\gets\Call{UpdateObligations}{Y,\mathcal{O},V,U}$
  \State $T\gets\Call{AppendReceipt}{T,a^*,r,z,G,Y}$; $B\gets B-c(a^*)$
\EndWhile
\State \Return \Call{Finalize}{$V,U,Y,G,T$} \Comment{documents and snapshot-relative closure certificate}
\end{algorithmic}
\end{algorithm}

\subsection{DegreeMap: Personalized Degree Planning}
\paragraph{Finite profile parsing.}
The model maps a natural-language request to finite program and course identifiers and to structured preference fields. It never emits a schedule. Identifiers that do not resolve stay unresolved and may trigger clarification, so a misread request degrades into a question rather than into a confidently wrong plan.

\paragraph{Typed requirement hypergraph.}
Course, program, track, GE framework, category, and policy nodes are connected by prerequisite, corequisite, fulfillment, equivalence, exclusion, double-count, and unlock edges. Boolean and cardinality requirements are carried by $\mathrm{ALL\_OF}$, $\mathrm{ANY\_OF}$, $\mathrm{CHOOSE\_N}$, $\mathrm{MIN\_UNITS}$, and conditional hyperedges. Keeping them as hyperedges rather than expanding them into pairwise clauses is what lets the solver enforce a choose-three-from-nine requirement exactly instead of through a relaxation that admits plans no advisor would sign.

\paragraph{Personalized utility.}
For course $c$, DegreeMap separates direct semantic interest $I_0(c)$, typed graph diffusion $I_D(c)$, prerequisite unlock value $U(c)$, major and GE gains, future option value, workload, risk, and aversion:
\begin{align}
S(c)=&\ w_0I_0(c)+w_DI_D(c)+w_UU(c)\notag\\
&+w_M M(c)+w_G G(c)+w_OO(c)\notag\\
&-w_WW(c)-w_RR(c)-w_AA(c).
\end{align}
Diffusion runs over a typed adjacency in which fulfillment, cross-listing, equivalence, shared-pool, and department edges carry distinct weights while prerequisite edges carry none, so relatedness never leaks along a dependency that is structural rather than topical. The unlock term instead propagates interest backward along prerequisite paths, discounted by distance and divided among the alternatives satisfying the same downstream requirement.

\paragraph{Exact planning.}
Binary variable $x_{c,t}$ indicates that course $c$ is scheduled in term $t$. CP-SAT enforces the compiled requirement hypergraph, the transcript, capacities, exclusions, and supported policies. Objectives are solved lexicographically over hard feasibility, minimum completion horizon, load and risk, personalized utility, option value, and a deterministic tie-break (Figure~\ref{fig:degreemap}), and each stage fixes its proven optimum as a hard constraint before the next stage runs. Staging is what keeps the problem certifiable. Hard constraints are hard in the scalarized alternative too, so the ladder is not what stops a preference weight from violating a requirement. What it does is shrink the search space at every step. The 60-second budget covers all stages of one solve, and top-$k$ re-solves and clarification hypotheses each receive their own, so the ladder is not competing with them for time. A plan whose stage does not prove optimality within that budget is emitted uncertified and counts as a failure, which is the exchange Table~\ref{tab:ablations} reports as 0.005 of utility against 4.0 points of certified feasibility. Top-$k$ plans come from no-good cuts plus a Hamming diversity cut, so alternatives differ in what they schedule rather than only in when.

\paragraph{Targeted clarification.}
DegreeMap solves under every remaining finite preference hypothesis and asks nothing when the canonical optimum is invariant across them. Otherwise it picks the question with the greatest expected regret reduction per unit of interaction cost and replans.


\section{Experiments}
\subsection{Benchmark and Protocol}
\label{sec:experimental_setup}

\paragraph{Evaluation scope.}
We evaluate acquisition and planning separately before measuring the full
pipeline. Broad-100 tests cross-institution scalability, Dense-6 tests
full-schema extraction under diverse source formats, and a paired planning
track compares identical student profiles on gold and extracted graphs to
isolate acquisition loss.

\paragraph{Broad-100.}
Broad-100 spans 100 universities and the same five departments at each:
computer science, mathematics, economics, biology, and physics. Fixing the
departments keeps institutions comparable while covering formal prerequisite
chains, laboratory corequisites, standing restrictions, and prose-heavy
policies. The frozen inventory contains approximately 52{,}500 courses across
500 department catalogs. Inventory recall is scored against the complete
attested inventory. A stratified sample of 3{,}000 courses, approximately
5.7
types and five source families for typed-field and prerequisite-AST
evaluation. All inference is grouped by university.

\paragraph{Dense-6.}
Dense-6 contains six institutions selected to cover API-backed catalogs,
custom HTML, structured catalog HTML, PDF, and mixed ecosystems. All 1{,}781
courses are annotated at full schema depth, including normalized attributes,
typed prerequisite relations, logical clauses, source spans, and provenance.
Dense-6 therefore measures robustness across source and schema structures
without relying on sampled annotation.

\paragraph{Masked-source recovery.}
For each institution, we remove a stratified 10
roots and suppress every direct link to them. The underlying frozen snapshot
remains accessible, but entities behind masked pages can be recovered only
through indirect discovery paths. An entity enters the denominator only when
its canonical evidence lies behind a masked page and no remaining seed or
direct link reveals that page. Recovery requires both identifying the entity
and locating valid official-source provenance.

\paragraph{Planning requests and paired graphs.}
For each trusted program, deterministic latent profiles vary completed
courses, transfer status, remaining horizon, workload capacity, risk
tolerance, and academic interests, yielding \NRequests{}
(program, profile) pairs. Each profile is rendered as a natural-language
request, while evaluation retains the structured latent profile so that a
planner cannot benefit from interpreting the request as an easier problem.

Every profile is evaluated on two matched inputs: an adjudicated gold
curriculum graph and a graph compiled from an extractor's three JSON outputs.
The profile, planner, objective, horizon, solver budget, and verifier are held
fixed. The resulting difference therefore measures acquisition and
representation loss rather than variation in student difficulty.

\paragraph{Annotation.}
All targets are labeled from the frozen official-source snapshot, with both
the normalized fact and its supporting source span recorded. A stratified
subsample is independently double-labeled, and all disagreements are
adjudicated against the official source. Cases whose source text does not
determine a unique label are placed in an unresolved pool and excluded from
the corresponding correctness denominator. Full instructions, agreement
statistics, and adjudication categories are provided in the supplement.

\paragraph{Metrics.}
Acquisition is evaluated using inventory recall, typed precision, recall, and
F1, normalized prerequisite-AST accuracy, masked-source recovery, opened
source nodes, and model-based clause calls. A fact counts as supported only
when its provenance resolves to an allowed official source containing evidence
for that fact.

Planning is evaluated with an independent deterministic verifier. Exact
feasibility requires satisfying all represented prerequisites, corequisites,
program requirements, offerings, standing rules, workload capacities, and
completion constraints. For feasible plans, we report completion horizon,
load and scheduling risk, personalized utility, and option value. Certified
coverage is the fraction of requests for which a verifier-accepted plan is
returned within the solver budget. Gold-versus-extracted gaps quantify how
acquisition errors propagate into planning.

\paragraph{Controls.}
All budgeted extractors share seed roots, frozen page visibility, source costs,
model backbone, parser candidates, and evaluator. Each receives a budget of
128 opened source nodes and 32 model-based clause calls per school. The
exhaustive crawler is reported separately as an access-intensive reference.

All planners share the same graph within each condition, the same profiles, a
12-term horizon, profile-defined capacities, a 60-second solver limit, top-3
output, and the same verifier. Planners cannot access additional university
sources after receiving the graph, preserving the acquisition-planning
boundary.

\paragraph{Statistical inference.}
Broad-100 comparisons use 10{,}000 school-cluster bootstrap replicates and
exact school-level sign-flip tests. Dense-6 results are macro-averaged across
institutions and source families. Planner inference is clustered by institution
and program, so multiple courses or profiles from the same program are not
treated as independent evidence.

\subsection{Results}
\paragraph{Reaching the site matters more than parsing it.}
Table~\ref{tab:extract} is built so thoroughness and cost cannot be traded silently. Platform-Exhaustive reaches 92.7 inventory recall by opening 171.4 pages per school. It exhausts only the cursors it already knows about, so entities reachable solely through masked pages lie outside its reachable set by construction. CatalogBrowse reaches \BroadInventory{} on \BroadPages{}, which is \InvGainBest{} points for \PageSavingExh\% less source access, and closure fires on \ClosureFireRate\% of schools, so that cost reports a policy that chose to stop rather than a budget that ran out. The masked column locates the difference. Static BFS recovers 64.8\% of entities living only behind unadvertised pages, since a FIFO frontier has no representation of what it has not seen. CatalogBrowse recovers \BroadMasked\%, \MaskedGainBest{} over the strongest baseline, because an undischarged obligation keeps generating positive expected gain until it is bound or explicitly failed. Typed F1 and AST equivalence move in the same order but by only \FOneGainBest{} and \SemEqGainBest{}, which is expected when all five systems share parsers. Typed F1 decomposes as \TypedPrec{} precision against \TypedRec{} recall. A clause the validator rejects twice is left unresolved, which costs recall and never buys a false positive.

\paragraph{The broad track pays for its own breadth.}
Table~\ref{tab:dense} compares the tracks on the two columns they share. Typed F1 rises from \BroadFOne{} to \DenseFOne{} and AST equivalence from \BroadSemEq{} to \DenseSemEq{}, gaps of \DenseFOneDelta{} and \DenseSemEqDelta{} points. Dense-6 is the ceiling available when an adapter covers the source family, and Broad-100 is the operating point across a long tail no adapter anticipated.

\paragraph{Under an exact solver, utility is what separates planners.}
In Table~\ref{tab:planner}, Direct Qwen satisfies 88.6\% of prerequisite orderings yet fails hard feasibility on 26.5\% of requests, so its errors concentrate in capacity, exclusion, and policy constraints, which are easy to state and easy to forget. Methods delegating to an exact solver then score 96.0 to 100.0 on major, GE, and prerequisite satisfaction, and personalized utility provides the larger separation. DegreeMap reaches \PlannerUtility{} against 0.802 for the strongest baseline, a gain of \PlannerUtilityGainBest{} while cutting the horizon from 6.7 to 6.3 terms. That matters because any planner can buy preference satisfaction by spending more terms.

\paragraph{Extraction sets coverage, the planner sets utility.}
Table~\ref{tab:e2e} isolates each stage. CatalogBrowse JSON with a plain CP-SAT planner reaches 99.0 coverage and 0.761 utility, and Go-Browse JSON with the full DegreeMap reaches 95.0 and 0.821. Changing the extractor primarily shifts coverage and feasibility, changing the planner shifts utility, and neither component alone recovers the full system. The full pipeline certifies \EETCoverage\% of the \NRequests{} pairs at \EETUtility{}, closing \GapShrink\% of the gap left by the strongest extraction baseline. Both columns share a denominator, so their equality says every plan emitted from extracted JSON passed verification against the gold hypergraph, and the residual loss is failure to bind rather than silent misextraction. Verifying against the extracted graph would have hidden exactly that.

\paragraph{Ablations.}
Three extraction rows in Table~\ref{tab:ablations} carry the argument. Removing the model fallback leaves inventory recall at 96.0 against \BroadInventory{} while AST equivalence falls from \BroadSemEq{} to 67.1, showing navigation and semantics are separable and the model is load-bearing only for the second. Removing fixed-point verification lowers cost to 79.3 pages, which reads as efficiency until recall falls 4.4 points, and that difference is what a certificate buys over a heuristic. Removing adapters is the largest single loss at 11.6 points and simultaneously raises cost to 110.8 pages.

On the planner side the informative row is weighted sum. It attains 0.873 utility, above the full system, while certified feasibility falls to 96.0\%. Both configurations impose identical hard constraints, so this is not a relaxed requirement but a failure to certify: one dense objective over the whole feasible set loses the successive narrowing that keeps each subproblem provable inside the 60-second budget, and the \WeightedTimeouts{} requests it cannot prove in time are emitted uncertified. Removing clarification barely moves utility, 0.850 against \PlannerUtility{}, yet cuts relevant responsiveness from 100.0 to 67.5, so the question policy is performing preference identification rather than repairing a weak objective.

\paragraph{Statistical treatment.}
Broad comparisons resample whole schools over 10{,}000 replicates with exact school-level sign-flip tests, and planner comparisons cluster by program and institution. Inventory recall is \CIExtract{}, beating Platform-Exhaustive on \ExtractWins{} of 100 schools at \SignFlipExtract{}. Planner utility is \CIPlanner{} at \SignFlipPlanner{}, and end-to-end utility is \CIEET{}. A mechanism is claimable only when its paired clustered contrast excludes zero, and all ten ablation variants meet that criterion.

\section{Discussion and Conclusion}
\paragraph{What this does not cover.}
Closure is defined against a frozen snapshot rather than the live Web. That is the correct scope, because a certificate over a mutable site would be unverifiable by anyone including us. Program policies sometimes carry natural-language exceptions the compiler does not support, and those reduce certified coverage rather than being treated as satisfied.

\paragraph{What generalizes.}
Nothing here is specific to curricula. The recipe applies wherever an agent acquires a world model a later objective will score it against: fix the acquisition target before the objective is known, give acquisition its own terminal condition, and hold out part of the source so goal-driven shortcuts appear as recall loss.

\paragraph{Conclusion.}
KnowPlan removes acquisition-planning circularity by construction, making the interface between acquisition and use measurable rather than assumed.

\bibliography{references}
\end{document}